\documentclass[conference,10pt]{IEEEtran}
\IEEEoverridecommandlockouts
\usepackage{cite}
\usepackage{amsmath,amssymb,amsfonts}
\usepackage{algorithmic}
\usepackage{graphicx}
\usepackage{textcomp}
\usepackage{xcolor}
\usepackage{hyperref}
\usepackage{bigstrut}
\usepackage{makecell}
\usepackage{tabu}
\usepackage{color}
\usepackage{tabularx}

\newcommand{\uav}{Unmanned Aerial Vehicle}
\newcommand{\UAV}{UAV}

\newcommand{\bferra}{Bruno Ferrarini}
\newcommand{\bferraEM}{bferra@essex.ac.uk}

\newcommand{\sheh}{Shoaib Ehsan}
\newcommand{\shehEM}{sehsan@essex.ac.uk}

\newcommand{\kdm}{Klaus D. McDonald-Maier}
\newcommand{\kdmEM}{kdm@essex.ac.uk}

\newcommand{\maria}{Maria Waheed}
\newcommand{\mariaEM}{maria.waheed97@gmail.com}

\newcommand{\sania}{Sania Waheed}
\newcommand{\saniaEM}{saniawaheed97@gmail.com}

\newcommand{\mm}{Michael Milford}
\newcommand{\mmEM}{michael.milford@qut.edu.au}

\newcommand{\lag}{Lagout}
\newcommand{\cor}{Corvin}
\newcommand{\city}{Old-City}

\newcommand{\dref}{$I_{REF}$}

\newcommand{\dtrain}{$I_{r}$}

\newcommand{\vase}{VASE-JBL}

\def\BibTeX{{\rm B\kern-.05em{\sc i\kern-.025em b}\kern-.08em
    T\kern-.1667em\lower.7ex\hbox{E}\kern-.125emX}}
\begin{document}

\title{Visual Place Recognition for Aerial Robotics: Exploring Accuracy-Computation Trade-off for Local Image Descriptors\\
\thanks{This work has been supported by the UK Engineering and Physical Sciences
Research Council EPSRC [EP/K004638/1, EP/R02572X/1 and EP/P017487/1]}
}

\author{\IEEEauthorblockN{1\textsuperscript{st} \bferra{}}
\IEEEauthorblockA{\textit{CSEE School} \\
\textit{University of Essex}\\
Colchester, CO4 3SQ, UK \\
\bferraEM{}}
\and
\IEEEauthorblockN{2\textsuperscript{nd} \maria{}}
\IEEEauthorblockA{\textit{}\\
\textit{\quad COMSATS University} \\
\quad Islamabad, Pakistan \\
\quad \mariaEM{}}
\and
\IEEEauthorblockN{3\textsuperscript{rd} \sania{}\qquad \quad}
\IEEEauthorblockA{\textit{National University of\qquad} \\
\textit{Sciences and Technology (NUST) \qquad }\\
Islamabad, Pakistan \qquad \\
\saniaEM{}\qquad}
\and
\IEEEauthorblockN{\qquad \quad 4\textsuperscript{th} \sheh{}}
\IEEEauthorblockA{\qquad \quad \textit{CSEE School} \\
\textit{\qquad \quad University of Essex }\\
\qquad \quad Colchester, CO4 3SQ, UK  \\
\qquad  \quad \shehEM{}}
\and
\IEEEauthorblockN{\quad 5\textsuperscript{th} \mm{}}
\IEEEauthorblockA{\quad \textit{Science and Engineering Faculty} \\
\quad \textit{Queensland University of Technology}\\
\quad Brisbane, Australia \\
\mmEM{}}
\and
\IEEEauthorblockN{6\textsuperscript{th} \kdm{}}
\IEEEauthorblockA{\textit{CSEE School} \\
\textit{University of Essex}\\
Colchester, CO4 3SQ, UK \\
\kdmEM{}}
}

\maketitle

\begin{abstract}
Visual Place Recognition (VPR) is a fundamental yet challenging task for small \uav{} (\UAV{}). The core reasons are the extreme viewpoint changes, and limited computational power onboard a UAV which restricts the applicability of robust but computation intensive state-of-the-art VPR methods. In this context, a viable approach is to use local image descriptors for performing VPR as these can be computed relatively efficiently without the need of any special hardware, such as a GPU. However, the choice of a local feature descriptor is not trivial and calls for a detailed investigation as there is a trade-off between VPR accuracy and the required computational effort. To fill this research gap, this paper examines the performance of several state-of-the-art local feature descriptors, both from accuracy and computational perspectives, specifically for VPR application utilizing standard aerial datasets. The presented results confirm that a trade-off between accuracy and computational effort is inevitable while executing VPR
on resource-constrained hardware.
\end{abstract}


\section{Introduction}

Autonomous navigation of \UAV{}s has been receiving great attention recently \cite{mozaffari2016efficient,maffra2019real,villa2016overview}  as it has a wide variety of industrial applications, such as aerial imaging and surveying \cite{ascending}. As part of autonomous navigation, place recognition is critical for \UAV{} localization \cite{li2016real}. When the estimation of a vehicle position drifts due to accumulated errors over time, re-localization is possible when a reference location/already-visited place is detected \cite{wheeler2018relative}. Place recognition for a \UAV{} is usually addressed using visual information, hence called visual place recognition. This is motivated by the availability, cost, size and weight of modern cameras, which make them feasible to be installed even on a small aerial vehicle.  

VPR is a particularly challenging problem for a small \UAV{} due to the extreme viewpoint changes and limited computational power onboard \cite{maffra2019real}. 
State-ot-the-art VPR methods are only robust to small viewpoint changes \cite{milford2012seqslam,sunderhauf2013we} or their computation intensive nature \cite{chen2017only} makes their use prohibitive for small \UAV{}s
Some recent works \cite{maffra2018tolerant,maffra2019real} proposed VPR pipelines based on local image features as they can be extracted efficiently on resource-constrained hardware. As noted in \cite{maffra2018tolerant}, the choice of the local feature descriptor implies a trade-off between the computation efficiency and the accuracy of the overall visual place recognition system. This calls for detailed investigation into accuracy-computation trade-off for local feature descriptors specifically for VPR application.

To this end, this paper explores the accuracy-computation trade-off of several state-of-the-art local feature descriptors 
for VPR under mild to extreme viewpoint changes in small \UAV{}s using standard ground-aerial image datasets \cite{maffra2019real}. Precision-Recall and computation time are used as the basis of exploring this accuracy-computation trade-off. Results are presented for SIFT \cite{lowe2004distinctive}, SURF \cite{baya_speeded-up_2008}, BRISK\cite{leutenegger2011brisk}, AKAZE\cite{alcantarilla2011fast} and ORB\cite{rublee2011orb} descriptors. However, the proposed evaluation method can be extended to any local image feature descriptor. The results presented confirm that the more computationally expensive descriptors yield to better accuracy at the cost of a longer localization time. A trade-off between accuracy and computation effort appears inevitable while executing VPR on a limited resource hardware.

The rest of this paper is organized as follows. Section \ref{sec:work} provides an overview of related work. Section \ref{sec:experiment} describes the method and criteria used for performance evaluation. The experimental results are presented in Section \ref{sec:res}. Finally, conclusions are given in Section \ref{sec:conclusions}.

\section{Related Work}
\label{sec:work}


Visual place recognition is critical for developing a practical and self-reliant UAV that does not require an external tracking or positioning system \cite{maffra2017loop}. The fundamental task for VPR is building and searching an image database to determine if a previously visited location is detected. The goal is to implant this ability in a UAV for it to be able to re-localize itself. However, this is challenging due to the extreme viewpoint changes experienced by a UAV and limited computational power on-board \cite{zaffar2019state}.  Several state-of-the-art VPR methods have been rendered unfeasible as they fail to perform satisfactorily under such conditions \cite{maffra2019real}. Convolutional Neural Networks (CNNs) display high performance for visual place recognition but are computationally intensive to be a suitable choice for a small UAV \cite{arandjelovic2016netvlad}. 
\begin{figure}[t]
\centering

\includegraphics[width=8.5cm]{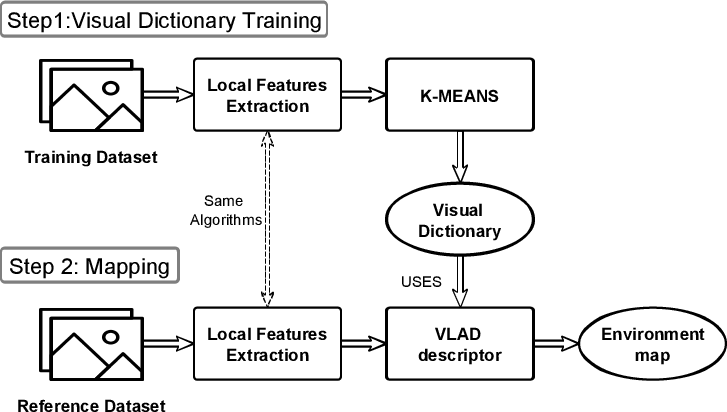}
\caption{Mapping occurs in two steps. First a visual dictionary to quantize the feature space is computed with k-means algorithm. In the second step, a map consisting of VLAD descriptors is built using the visual dictionary from the first step.
}
\label{fig:mapping}
\end{figure}
On the other hand, local feature descriptors exhibit higher potential for usage in a UAV as they are computationally less expensive. Since there is a computation-accuracy trade-off involved in the choice of a feature descriptor for UAV based visual place recognition, the selection process requires detailed analysis. To cope with the low computational power on-board a UAV, several other approaches have also been employed including the Bag of Words (BoW) approach \cite{sivic2003video}. This method can increase efficiency by creating visual dictionaries, by taking into account locally invariant feature descriptors, to match the query image vocabulary. Currently, the BoW approach tries to deal with pose change by utilizing feature descriptors, such as SIFT\cite{lowe2004distinctive} and SURF\cite{baya_speeded-up_2008} but they often fail when encountered with extreme viewpoint changes. As a result, use of range sensors \cite{dube2017segmatch} and structural descriptors \cite{cieslewski2016point} have received attention as they offer better performance to larger viewpoint changes. However, they involve higher power consumption rendering their use unfeasible for a small UAV. Other approaches undertaken include the use of FABMAP \cite{cummins2008fab} as it alleviated the BoW performance but its applicability in a UAV is restricted by the high variance in performance to even small viewpoint change. Another methodology is using binary feature detectors such as ORB \cite{rublee2011orb}, BRISK \cite{leutenegger2011brisk} and  FREAK \cite{alahi2012freak} that depicted performance similar to far more computationally expensive features, like SIFT or SURF. 
With multiple possibilities, each with its perks and shortcomings, it is a tedious job to rank or select any one of these methods unbiasedly. The trade-off between VPR accuracy and the required computational effort makes the selection procedure a strenuous task. This paper proposes to explore this trade off and evaluate performance difference while avoiding any inclination towards a particular image feature descriptor by selecting a dataset comprising of both ground and aerial images.

\begin{figure}[!tb]
\centering

\includegraphics[width=8.5cm]{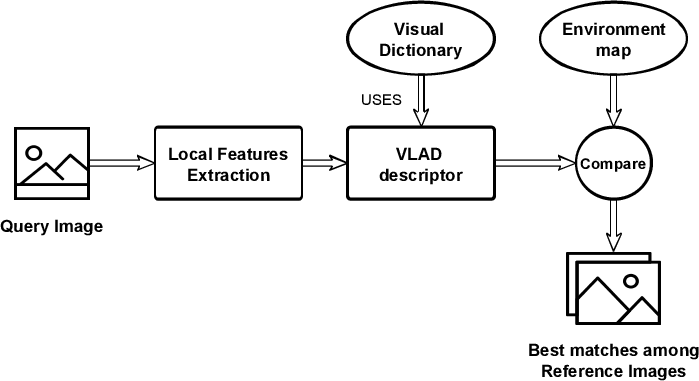}
\caption{The VLAD descriptor of a query image is computed and compared with those forming the environment map. Reference images  are ranked by similarity between their descriptor and query image's descriptor and returned as a result.
}
\label{fig:localiz}
\end{figure}

\section{Experiment Setup}
\label{sec:experiment}


The proposed approach evaluates local image feature descriptors using VPR in the form of an image retrieval task. Local image features descriptors are used to build a map of the environment from a set of reference images showing previously visited places. During the localization phase, the map is searched to retrieve the reference images that match with a query image (i.e. a frame captured by the onboard camera). Localization succeeds if the query frame is correctly matched with the key frames in the map corresponding to the current position within the environment. 
The VPR algorithm and the evaluation criteria used to assess local image feature descriptors are detailed as follows.

\subsection{Mapping}

A  map represents the knowledge about an environment. In our setup, a map is built from a set of images showing environment locations denoted as reference dataset, \dref{}. 
%
For each image in \dref{}, a set of  local image features descriptors are computed and then combined into a compact image descriptor using  VLAD\cite{jegou2010aggregating,arandjelovic2013all}. The resulting set of VLAD descriptors are finally organized in a ball-tree structure \cite{omohundro1989five} in order to make the localization faster.
VLAD requires a visual dictionary to quantize the feature set to combine in a single image descriptor. The visual dictionary, $V_k$, is computed with k-means clustering \cite{hartigan1979algorithm} using the features extracted from a training dataset, \dtrain{}. The local feature descriptor used to obtain the features set to train $V_k$  is the same used for computing the VLAD descriptors. Figure \ref{fig:mapping} summarizes the mapping stage.
%

\subsection{Localization} 

Localization consists of finding the best matching images in \dref{} for a query image. The image retrieval process is illustrated in the diagram in Figure \ref{fig:localiz}. A VLAD descriptor is computed for the query image and compared with the VLAD descriptors of the mapped reference images. The reference images are ranked for their similarity with the input image and the resulting sequence is returned as a result of the query. Highest rank means highest similarity between a reference image and the query image. If the localization succeeds, the images ranked at top show the same place as the query image. 

\subsection{Evaluation}

The highest ranked images extracted from the mapping dataset should correspond to the same place of the query image. The accuracy is evaluated with Precision and Recall (PR) while the metric for the efficiency is the required time to complete a query. In particular, the execution time includes the time spent to compute an image descriptor for the query image and the time for searching the map.

\section{Results}
\label{sec:res}

\begin{figure}[!tb]
\centering

\includegraphics[width=8.5cm]{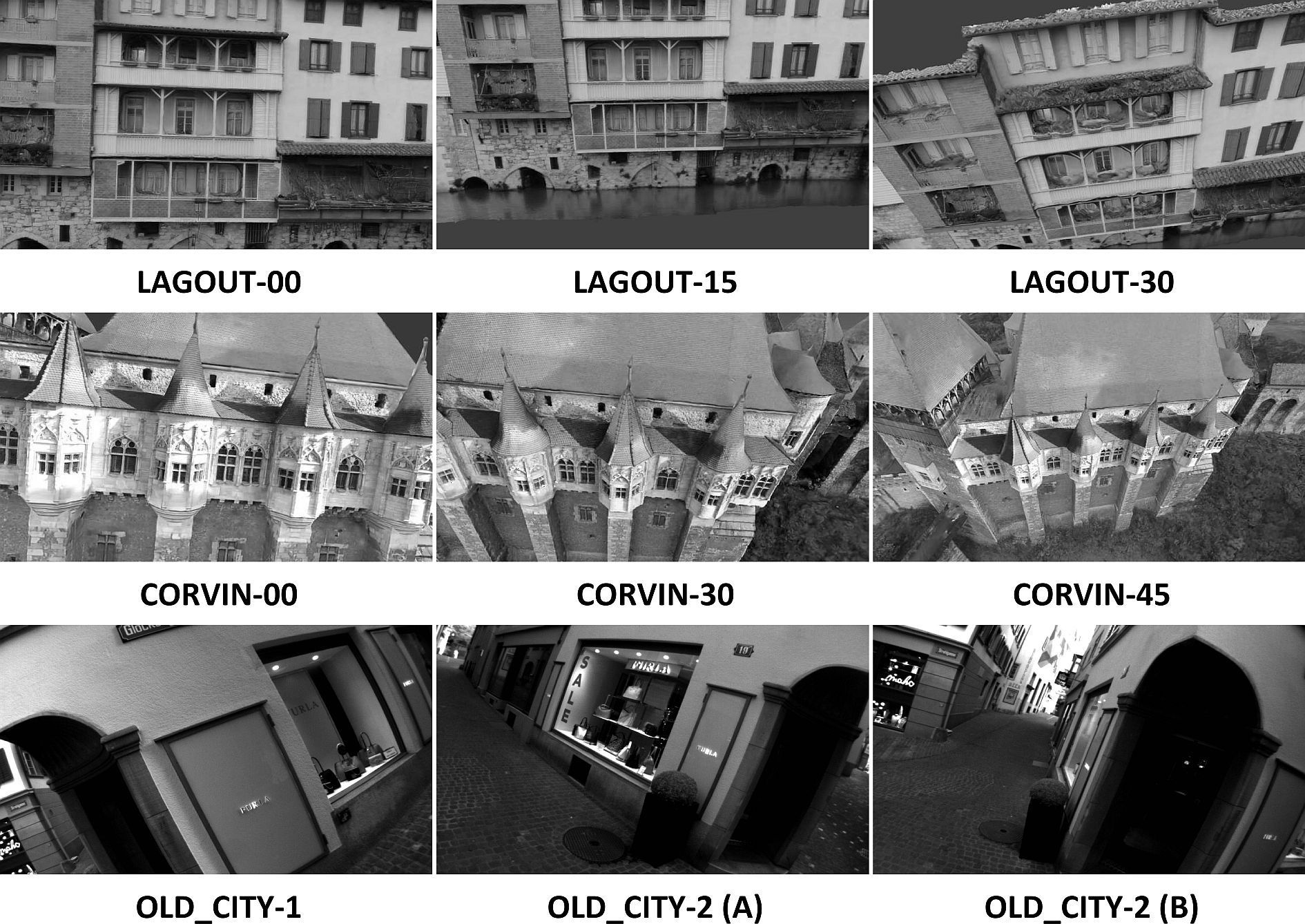}
\caption{A place from each dataset is shown as it appears in different loops.
}
\label{fig:samples}
\end{figure}

The descriptors considered for the tests are SIFT\cite{lowe2004distinctive}, SURF\cite{baya_speeded-up_2008}, BRISK\cite{leutenegger2011brisk}, AKAZE\cite{alcantarilla2011fast} and ORB\cite{rublee2011orb}. Each descriptor has been used with its native local feature detector stage as described in the original papers.
The datasets used for the tests are \lag{}, \city{} and \cor{} datasets \cite{maffra2019real}, which include both ground and aerial images scenes captured from a wide variety of viewing angles.
\lag{}
is a synthetic dataset consisting 
of 4 aerial footage captured at different angles: $0^\circ$, $15^\circ$, $30^\circ$ and $45^\circ$.  \cor{} is similar to \lag{} and includes  three aerial loops around the \cor{} Castle at $0^\circ$, $30^\circ$ and $45^\circ$. \city{} consists of two long loops captured in an urban environment which present a wide range of views of the same locations. Figure \ref{fig:samples} provides a sample from those datasets.
%
\begin{figure}[!t]
\centering

\includegraphics[width=8.5cm]{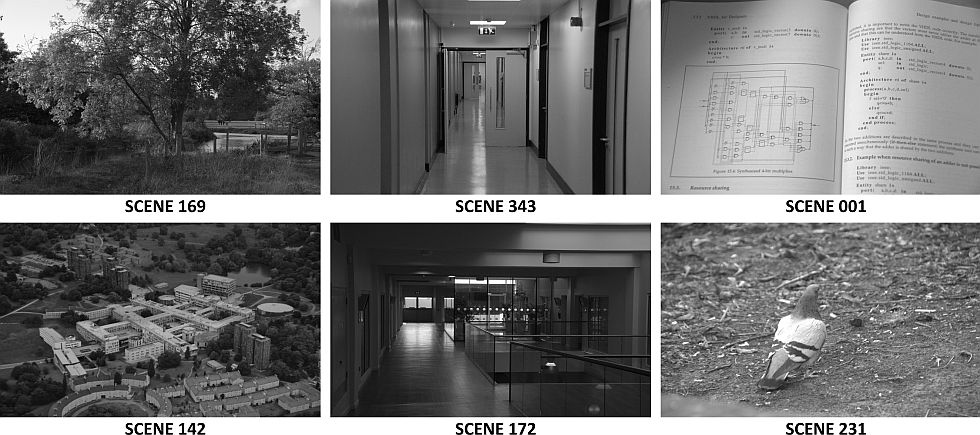}
\caption{Scene images from \vase{} dataset.
}
\label{fig:vase}
\end{figure}
The VPR algorithm used for the tests has been implemented on top of the source code available at \cite{pyVLAD} while for the local feature descriptors has been used the OpenCV \cite{itseez2015opencv} implementation. The reference dataset used for mapping are the loop at $0^{\circ}$ for \lag{} and \cor{} datasets and \city{}-1 for \city{} \footnote{\url{http://www.v4rl.ethz.ch/research/datasets-code/v4rl-wide-baseline-place-recognition-dataset.html}}.
The visual dictionary has been trained using a datasets which is not related with any of \lag{}, \cor{} and \city{} datasets. The training data consited of an image for each of the scenes available in \vase{} dataset \cite{ehsan2012jpeg} for a total of 539 images showing a wide variety of outdoor and indoor scenes captured in real-world environments (Figure \ref{fig:vase}). 
The descriptors has been ran with the default parameter as they are set in the release 3.4.2.17 of OpenCV.
The size of the visual dictionary has a significant impact on the VPR's accuracy. In order to maximize the VPR's accuracy with each descriptor, the visual dictionary length has been set at 2048 words for SURF and SIFT, 1024 for BRISK and AKAZE and 256 words for ORB.
%
%
\subsection{Accuracy and Computation Time}

\begin{figure*}[!th]
\centering

\includegraphics[width=17cm]{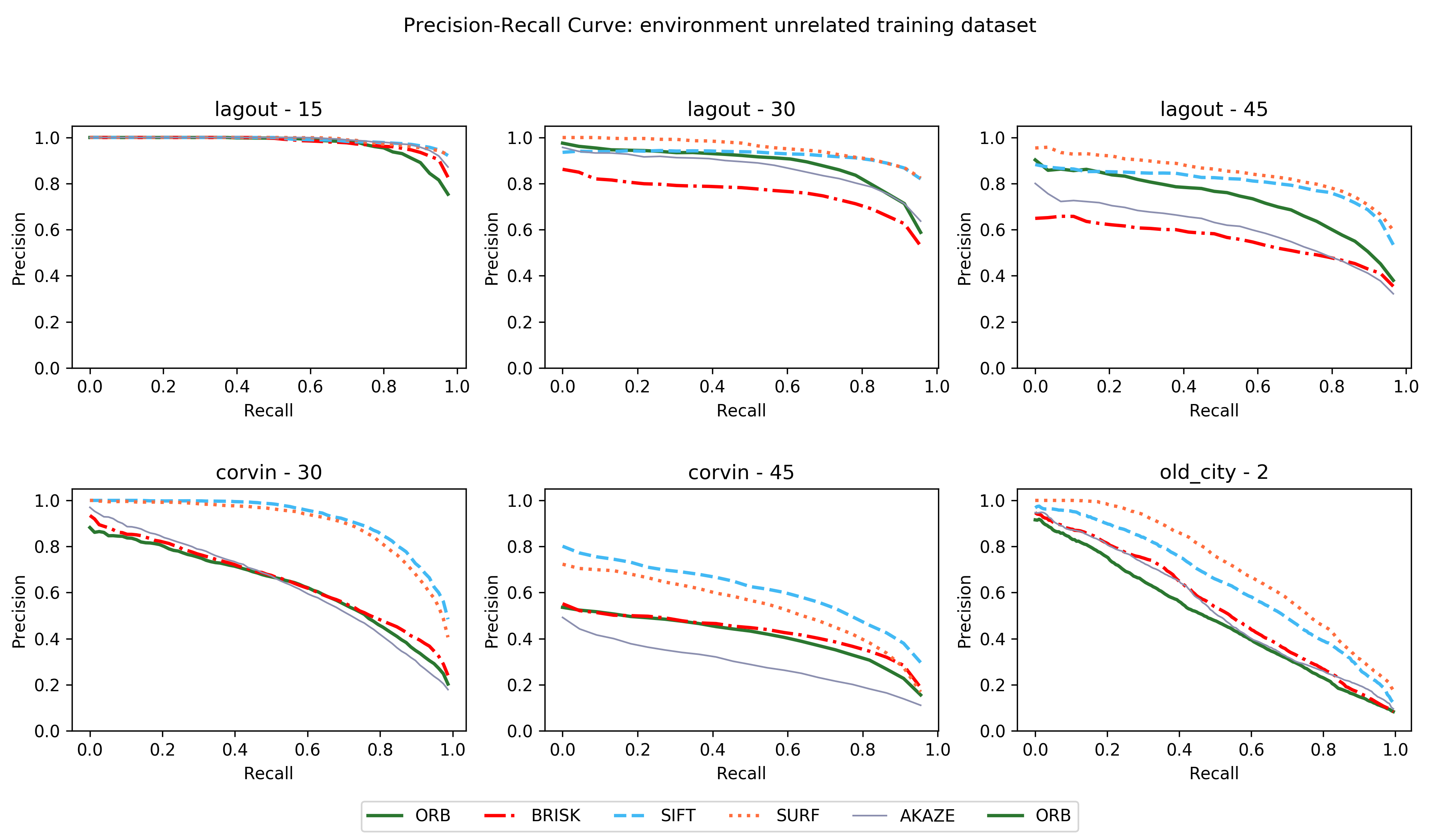}
\caption{Precision-Recall curves for ORB, BRISK, SIFT, SURF and AKAZE with \lag{}, \cor{} and \city{} datasets.}
\label{img:PR}
\end{figure*}

\begin{figure}[!b]
\centering

\includegraphics[width=8.7cm]{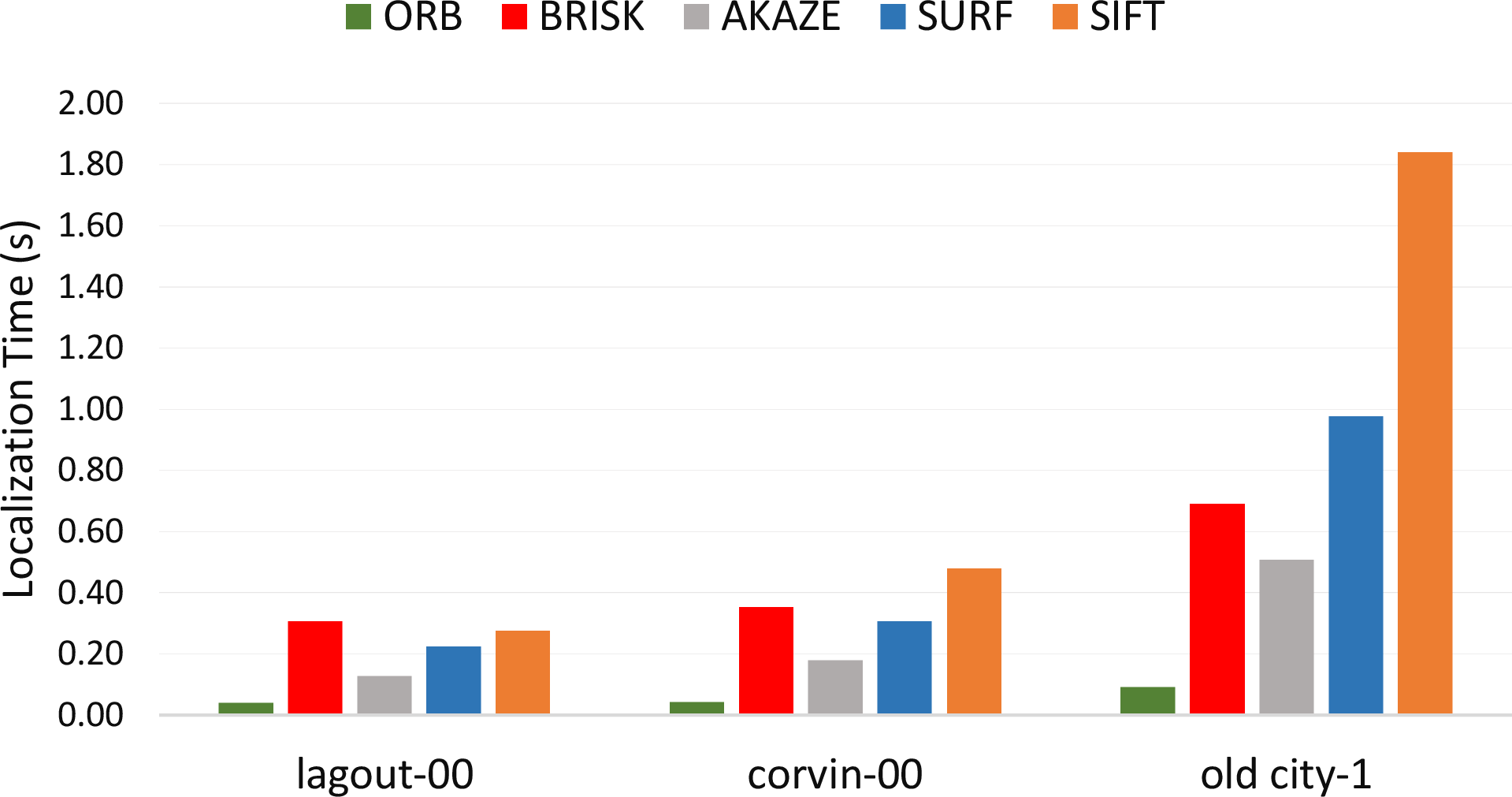}
\caption{The time required for localization in \lag{}, \cor{} and \city{} environments using an Intel i7-7700K CPU.}
\label{img:ET}
\end{figure}

Figure \ref{img:PR} shows the Precision-Recall curves (PR curves) for each of the assessed descriptors. The VPR exhibits the highest performance when uses SURF features with the only exception of \cor{}-45 and \cor{}-30 where SIFT outperform SURF by a close gap. 
Figure \ref{img:ET} shows the time required for localization for each descriptor and dataset recorded 
with a single thread on an Intel i7-7700K CPU.
The execution is longer on \city{} because the larger number of images included in its map: there are 6711 images in \city{}-1 while only 1183 and 336 images in \cor{}-00 and \lag{}-00 respectively.

While SIFT could be an alternative to SURF in terms of accuracy, a VPR system based on SURF is considerably more efficient. This gap is particularly wide when the localization occurs in \city{}, where SURF allows to complete the operation in about the 60\% of the time required when SIFT features are used. Although SURF based VPR can be a good trade-off in several cases, it still requires about $1s$ to complete the localization in \city{} using a desktop CPU. For \UAV{}s, whose computation power is more limited,  ORB can be a better option. While ORB based VPR is less accurate than SIFT and SURF based ones, it can complete the localization query 10 to 20 times faster.


\subsection{Environment-Related Training Data}

The results presented above are for a scenario where the VPR system is agnostic to the operating environment. In particular, the images used to train the visual dictionary is unrelated with \lag{}, \cor{} and \city{} datasets. This section proposes the results obtained for  visual dictionaries trained with the same data used to map the environment is order to provide the VPR system with some prior knowledge of the operating environment.  In details, to assess the localization performance with \lag{}, the reference loop \lag{}-00 has been used for both the mapping and training steps. The same setup has been used for \cor{} and \city{} datasets where the reference and training data are \cor{}-00 and \city{}-1 respectively. 

\begin{figure*}[!th]
\centering

\includegraphics[width=17cm]{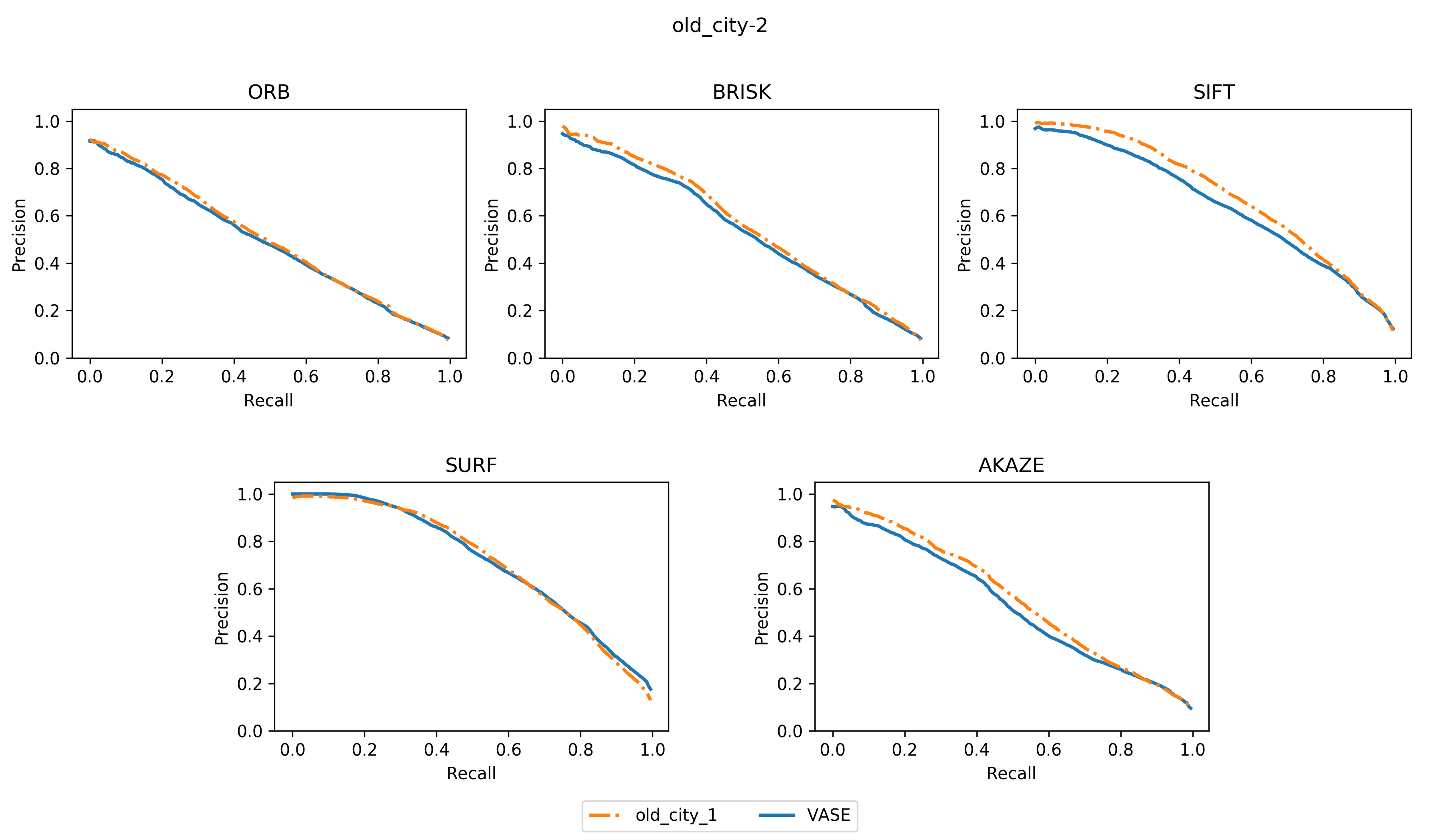}
\caption{A comparison between VPR trained with unrelated data from \vase{} dataset (blue curve) and with the same images used for the mapping (orange dot-dashed curve). The dataset used for the test is \city{}-2.}
\label{fig:KvU}
\end{figure*}

The results confirm that  the accuracy does not change significantly with respect to environment agnostic case. Figure \ref{fig:KvU} shows a comparison between the PR curves computed for the related and unrelated scenarios for \city{} dataset.  None of the assessed descriptors gain a significant improvement by using environment related data to train the visual dictionary. 
The reason lies in the low correlation between the local features of the mapping and the test loops.
Table \ref{tab:corr} show the correlation coefficients from every pair of training an mapping datasets used for the experiments. 
The difference between the correlation coefficients of \vase{} and the other training dataset is neglectable with every local feature descriptor. 
For example, \vase{} on \city{}-1 are unrelated datasets and, as expected, the related correlation coefficient for AKAZE features is small ($0.216$).  Regardless \city{}-2 and \city{}-1 show the same places, their AKAZE features can be considered unrelated as the related coefficient is just $0.214$, which is very close to the value computed for \vase{}. Consequently, there is not a significant advantage in using images from the environment with respect to random images to train the VPR algorithm used for the tests as not any significant correlation exists between features from similar datasets.

%
%

%

\section{Conclusions}
\label{sec:conclusions}


%

\begin{table}[!b]
  \centering
  \caption{Correlation coefficients between training and mapping datasets.}
\begingroup
\setlength{\tabcolsep}{4pt} 
\renewcommand{\arraystretch}{1.2} 
\begin{tabular}{|c|c|c|c|c|c|c|}
    \hline
    \makecell{Training \\ Dataset} & \makecell{Mapping \\ Dataset} & \rule{0pt}{4.0ex} ORB  \rule{0pt}{4.0ex}& BRISK & SIFT & SURF & AKAZE\\
    \hline
    \hline
    \lag{}-15 & \lag{}-00 & 0.151 & 0.216 & 0.308 & 0.223 & 0.215 \bigstrut[t]\\
    \lag{}-30 & \lag{}-00 & 0.154 & 0.221 & 0.312 & 0.217 & 0.218 \\
    \lag{}-45 & \lag{}-00 & 0.153 & 0.220 & 0.313 & 0.216 & 0.216 \\
    \vase{}  & \lag{}-00 & 0.148 & 0.220 & 0.310 & 0.222 & 0.215 \\
    \cor{}-30 & \cor{}-00 & 0.153 & 0.216 & 0.313 & 0.220 & 0.210 \\
    \cor{}-45 & \cor{}-00 & 0.149 & 0.219 & 0.310 & 0.218 & 0.213 \\
    \vase{}  & \cor{}-00 & 0.155 & 0.219 & 0.312 & 0.219 & 0.214 \\
    \city{}-1 & \city{}-1 & 0.152 & 0.222 & 0.310 & 0.22  & 0.214 \\
    \vase{}  & \city{}-1 & 0.152 & 0.222 & 0.307 & 0.222 & 0.216 \bigstrut[b]\\
    \hline
\end{tabular}
  \label{tab:corr}%
\endgroup

\end{table}%

This paper proposes a comparison of several  state-of-the-art local  local feature descriptors
for VPR under mild to extreme viewpoint changes in small \UAV{}s using ground-aerial image datasets. 
Localization accuracy is very important for VPR but it is not the only property to be considered in \UAV{}s. As \UAV{}s are vary agile vehicles, they need to re-localize quickly but, at the same time, they are often equipped with resource-constraint hardware. Driven by this consideration, the evaluation of local feature descriptors has been made on the basis of the accuracy and the computational effort to complete localization in order to determine the descriptor with the best trade-off for \UAV{} applications. 
The results show that the best accuracy can be reached using SURF and SIFT descriptors at the cost of long localization time. ORB can be a better option for \UAV{}s as is allow much faster localization while keeping a reasonable accuracy with most of the datasets considered for the experiments. 
%


\end{document}